\pdfoutput=1
\documentclass[10pt,twocolumn,letterpaper]{article}

\usepackage[pagenumbers]{cvpr} 

\usepackage{graphicx}
\usepackage{amsmath}
\usepackage{amssymb}
\usepackage{adjustbox}
\usepackage[accsupp]{axessibility}
\usepackage{mwe}
\usepackage{caption}
\usepackage{booktabs}
\usepackage{multirow}
\usepackage[table]{xcolor}
\usepackage{bm}
\usepackage{xcolor}
\usepackage{times}
\usepackage{epsfig}
\usepackage{comment}
\usepackage{stfloats}
\usepackage{url}
\newcolumntype{H}{>{\setbox0=\hbox\bgroup}c<{\egroup}@{}}
\makeatletter
\def\@fnsymbol#1{\ensuremath{\ifcase#1\or \dagger\or \ddagger\or
   \mathsection\or \mathparagraph\or \|\or **\or \dagger\dagger
   \or \ddagger\ddagger \else\@ctrerr\fi}}
\makeatother
\usepackage[capitalize]{cleveref}
\crefname{section}{Sec.}{Secs.}
\Crefname{section}{Section}{Sections}
\Crefname{table}{Table}{Tables}
\crefname{table}{Tab.}{Tabs.}

\def\cvprPaperID{226} 
\def\confName{CVPR}
\def\confYear{2022}

\begin{document}
\newcommand{\yueqi}[1]{{\color{orange}[Yueqi: #1]}}
\title{HyperDet3D: Learning a Scene-conditioned 3D Object Detector}

\author{
Yu Zheng\textsuperscript{\rm 1,3}
\quad Yueqi Duan\textsuperscript{\rm 2}\thanks{Corresponding author}
\quad Jiwen Lu\textsuperscript{\rm 1,3} 
\quad Jie Zhou\textsuperscript{\rm 1,3}
\quad Qi Tian \textsuperscript{\rm 4}\\
\textsuperscript{\rm 1}Department of Automation, Tsinghua University\\ 
\textsuperscript{\rm 2}Department of Electronic Engineering, Tsinghua University\\
\textsuperscript{\rm 3}Beijing National Research Center for Information Science and Technology\\
\textsuperscript{\rm 4}Huawei Cloud \& AI, China\\
{\tt\small zhengyu19@mails.tsinghua.edu.cn, \{duanyueqi,lujiwen,jzhou\}@tsinghua.edu.cn, tian.qi1@huawei.com}
}

\makeatletter
\let\@oldmaketitle\@maketitle
\renewcommand{\@maketitle}{\@oldmaketitle
\begin{minipage}{\textwidth}
\vspace{-9mm}
\centering
\includegraphics[width=.95\linewidth]{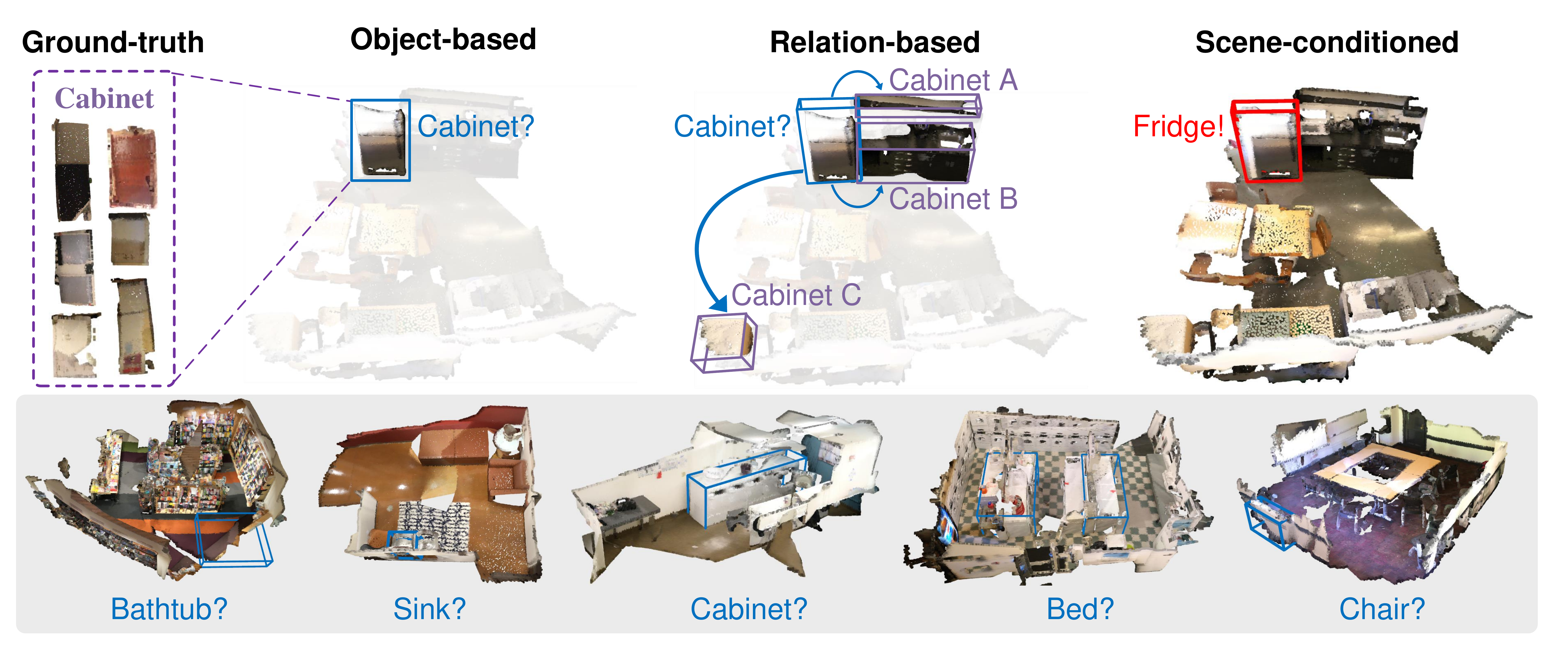}
\vspace{-3mm}
\captionof{figure}{
Exemplified predictions that highlight the importance of scene-conditioned knowledge. 
In the upper example, by observing the detection candidate in the object level, we can easily recognize it as \textit{cabinet} by comparing it with groundtruth cabinets, or relating it with other surrounding cabinets. 
However, conditioned on the prior knowledge that the object candidate lies in a kitchen-like scene, we may infer that it is a \textit{fridge}. 
We also illustrated 5 wrong detections which go against the scene-conditioned knowledge in the lower half, which are \textit{bathtub} in a library, \textit{sink} in an office, \textit{cabinet} or \textit{bed} in a laundry room, and \textit{chair} embedded in the wall of a meeting room. 
Note that point clouds are all colored only for easy illustration and not utilized in our method. 
(\textit{Best viewed in color.})
\label{fig:teaser}}
\end{minipage}
\bigskip}
\makeatother

\maketitle
\begin{abstract}
   A bathtub in a library, a sink in an office, a bed in a laundry room -- the counter-intuition suggests that scene provides important prior knowledge for 3D object detection, which instructs to eliminate the ambiguous detection of similar objects.
In this paper, we propose HyperDet3D to explore scene-conditioned prior knowledge for 3D object detection. 
Existing methods strive for better representation of local elements and their relations without scene-conditioned knowledge, which may cause ambiguity merely based on the understanding of individual points and object candidates. 
Instead, HyperDet3D simultaneously learns scene-agnostic embeddings and scene-specific knowledge through scene-conditioned hypernetworks. 
More specifically, our HyperDet3D not only explores the sharable abstracts from various 3D scenes, but also adapts the detector to the given scene at test time. 
We propose a discriminative Multi-head Scene-specific Attention (MSA) module to dynamically control the layer parameters of the detector conditioned on the fusion of scene-conditioned knowledge. 
Our HyperDet3D achieves state-of-the-art results on the 3D object detection benchmark of the ScanNet and SUN RGB-D datasets. 
Moreover, through cross-dataset evaluation, we show the acquired scene-conditioned prior knowledge still takes effect when facing 3D scenes with domain gap.

\end{abstract}

\section{Introduction}
3D object detection has gained much attention in recent years, which is fundamental for applications such as autonomous driving, robotic navigation and augmented reality. 
Early works adopt sliding window~\cite{song2016deep} or 2D prior~\cite{lahoud20172d} to locate objects from RGB-D data. 
However, the orderless and sparse characteristic of point cloud makes it hard to directly employ the recent advances in 2D detection. 
To tackle this, view-based methods~\cite{chen2017multi} project the points into multiple 2D planes and apply standard 2D detectors. 
Volumetric convolution-based methods~\cite{maturana2015voxnet,li20173d} split points into regular grids, which is feasible for 3D convolutions.

Different from the aforementioned view-based and volumetric convolution-based methods, PointNet++~\cite{qi2017pointnet++} focuses on the local geometries while elegantly consuming raw point cloud, and thus widely used as backbone network in 3D detectors. 
Built on the PointNet++ network, VoteNet~\cite{qi2019deep} yields outstanding results by regressing offset votes to object centers from seed coordinates and corresponding local features. 
Following works incorporate probabilistic voting~\cite{du2020spot}, multi-level contextual learning~\cite{xie2020mlcvnet,xie2021vote,duan2019structural} and self-attention based transformer~\cite{misra2021end,liu2021group,pan20213d} to further enhance the local representations. 
These methods underline the importance of exploiting object-based and relation-based representation of local elements, such as individual points, detection candidates and irregular local geometries in a given point scan. 

However, the attributes of similar objects are ambiguous if we only look at themselves or relations. 
In this paper, we discover that the scene-level information provides prior knowledge to eliminate such ambiguity. 
As shown in Figure~\ref{fig:teaser}, with the absence of scene-conditioned knowledge, inferring the object-level features or their relations is inadequate for detecting the object candidate, which may lead to counter-intuitive detection results in the aspect of scene-level understanding. 
To our best knowledge, the acquisition of such scene-level information among various scenes by 3D detectors is yet to be fully studied. 

To this end, we propose HyperDet3D for 3D object detection on point cloud which leverages hypernetwork-based structure. 
Compared with the existing methods that focus on point-wise or object-level representation, our HyperDet3D learns the scene-conditioned information as prior and incorporates such scene-level knowledge into network parameters, so that our 3D object detector is dynamically adjusted in accordance with different input scenes. 
Specifically, the scene-conditioned knowledge can be factorized into two levels: scene-agnostic and scene-specific information. 
For the \textbf{scene-agnostic} knowledge, we maintain a learnable embedding which is consumed by a hypernetwork and iteratively updated along with the parsing of various input scenes during training. 
Such sharable scene-agnostic knowledge generally abstracts the characteristics of training scenes and can be utilized by the detector at test time. 
Moreover, since conventional detectors maintain the same set of parameters when recognizing objects in different scenes, we propose to incorporate the \textbf{scene-specific} information which adapts the detector to the given scene at test time. 
To this end, we attentionally measure how well the current scene matches a general representation (or how much they differ) by using the specific input data as query. 
We simultaneously learn the two levels of scene-conditioned knowledge by proposing a Multi-head Scene-Conditioned Attention (MSA) module. 
The learned prior knowledge is aggregated with object candidate features by late fusion, therefore providing more powerful guidance to detect the objects. 
Extensive experiments on the widely used ScanNet~\cite{dai2017scannet} and SUN RGB-D~\cite{song2015sun} datasets demonstrate that our method surpasses state-of-the-art methods by an obvious margin. 
Moreover, through cross-dataset evaluation, we show the scene-conditioned prior knowledge acquired by our HyperDet3D still takes effect when faced with domain gap.

\section{Related Work}

\textbf{3D Object Detection for Point clouds: }
Since spatial information is better preserved in point cloud, most state-of-the-art approaches consume raw 3D coordinates as input~\cite{yang20203dssd,shi2020pv,zheng2021se,li2021lidar}. 
Early methods group point cloud into stacked 3D voxels~\cite{maturana2015voxnet,zhou2018voxelnet} to generate more structured data, or restricts the grouping operation within the ground plane to achieve real-time detection~\cite{lang2018pointpillars}. 
\begin{figure*}[tb]
\centering
\includegraphics[width=.99\textwidth]{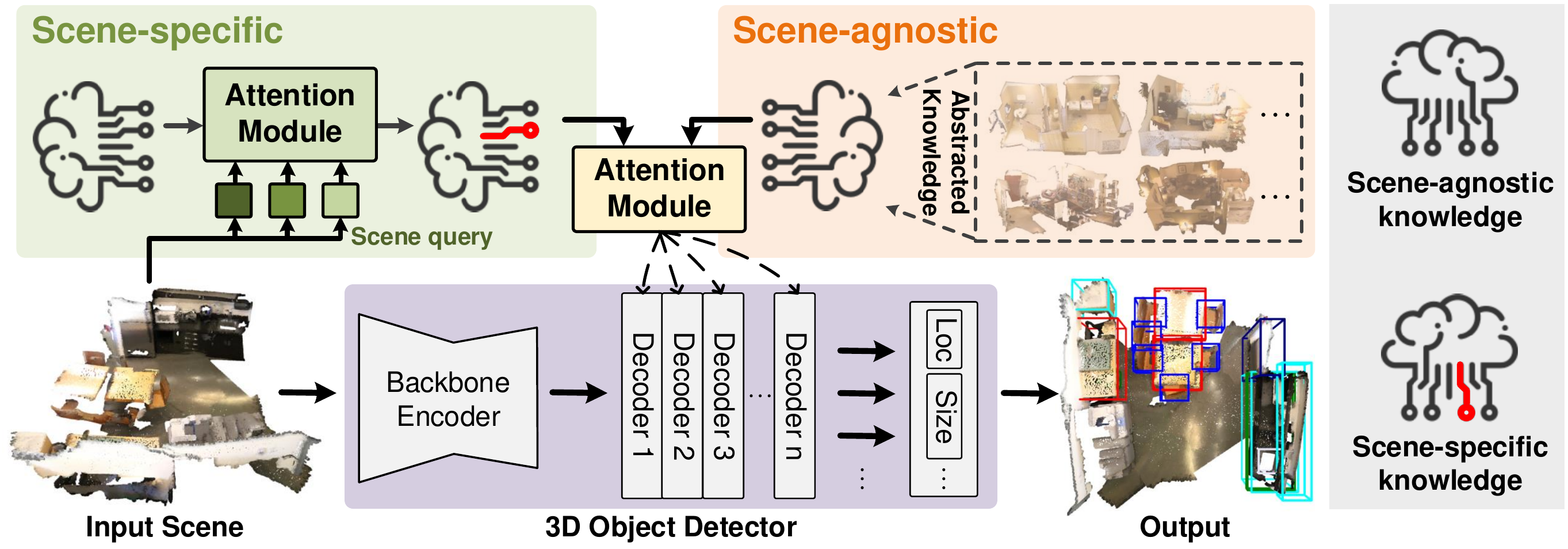}
\vspace{-3.5mm}
\caption{
Illustration of the proposed method. 
For a detection network in the lower half, our HyperDet3D in the upper half attentionally learns both scene-specific and scene-agnostic knowledge. 
Such scene-conditioned knowledge is then aggregated with object-level features in the decoder layers of detection network, so that the 3D detector is dynamically adjusted in accordance with different input scenes. 
The scene-agnostic knowledge is the sharable abstract learned from various scenes. 
The scene-specific knowledge attentionally measures how well a specific scene matches the general embedding (or how much they differ) by using the current scene as query. 
(\textit{Best viewed in color.})
} \label{fig:pipeline}
\end{figure*}
RCNN methods~\cite{shi2018pointrcnn,chen2019fastpoint,shi2020pv,li2021lidar} adopt PointNet-based~\cite{qi2017pointnet,qi2017pointnet++} module or use hybrid representation for better extracting and aggregating the point-wise feature. 
Inspired by the codebook learning in Hough Voting in 2D object detection~\cite{sun2010depth,gall2011hough}, VoteNet~\cite{qi2019deep} pioneerly construct the codebook of voting supervision from points to object centers by sampling and grouping proposed in PointNet++~\cite{qi2017pointnet++}. 
Based on the framework of VoteNet~\cite{qi2019deep}, H3DNet~\cite{zhang2020h3dnet} incorporates the votes to additional 3D primitives such as centers of box edges and surfaces. 
BRNet~\cite{cheng2021back} revisits the back-tracing operation in hough voting by querying the neighboring points around the object candidates. 
These methods enhance the feature representation of local elements by improving the voting mechanism itself. 
On the other hand, RGNet~\cite{feng2020relation} models the relation of object proposals by graph structures. 
SPOT~\cite{du2020spot} takes the probabilistic voting into account by measuring the information entropy of different local patches. 
MLCVNet~\cite{xie2020mlcvnet,xie2021vote} and PointFormer~\cite{pan20213d} incorporate multi-level attentional learning for object candidates and their contextual information. 
GroupFree3D~\cite{liu2021group} and 3DETR~\cite{misra2021end} introduce the classical Transformer~\cite{vaswani2017attention} architectures to the detection framework and achieve state-of-the-art performance. 
These methods explore the relation between local elements such as object candidates, local patches, point coordinates and their clusters. 

\textbf{HyperNetworks in Deep Neural Networks: }
HyperNetworks~\cite{ha2017hypernetworks} output the weights of the target network (called primary network) conditioned on specific input embedding. 
HyperNetworks have been embedded to replace the convolution or linear layers in image recognition~\cite{ha2017hypernetworks}, semantic segmentation~\cite{nirkin2021hyperseg}, neural architecture search~\cite{brock2017smash} and natural language modeling~\cite{ha2017hypernetworks}. 
In the field of 3D understanding, HyperCloud~\cite{spurek2020hypernetwork} and
HyperCube~\cite{proszewska2021hypercube} propose to produce a variety of shape representation for a single object by modifying the input to the hypernetwork. 
SDF-SRN~\cite{lin2020sdf} and MetaSDF~\cite{sitzmann2020metasdf} use hypernetworks to implicitly learn the object semantics within a category. 
More relevant to our work is HyperGrid~\cite{tay2021hypergrid} which designs the task-conditioned input embeddings of hypernetworks for a multi-task Transformer-based~\cite{vaswani2017attention} language model. 
Our HyperDet3D instead implicitly constructs the scene-specific and scene-agnostic embeddings for 3D object detection and, to our knowledge, is the first to incorporate hypernetworks in this task.

\section{Approach}
In this section, we first briefly introduce the overall architecture and some preliminaries. 
Next, we elaborate our proposed method. 
Finally, we provide the implementation details of the proposed method. 
\subsection{Overview and Preliminaries}
Figure~\ref{fig:pipeline} illustrates 3 key components in our proposed HyperDet3D, which are the backbone encoder, object decoder layer and detection head. 
Given an input point cloud $\mathbf{P}\in \mathbb{R}^{N\times 3}$, the backbone firstly downsamples the dense points into initial object candidates, as well as coarsely extracts their features through hierarchical architectures. 
For fair comparison, we consider PointNet++~\cite{qi2017pointnet++} as the backbone network similar to previous works~\cite{qi2019deep,zhang2020h3dnet,liu2021group}, which uses furthest point sampling (FPS) to uniformly cover the 3D space.  
Then the object decoder layers refine the candidate features by incorporating scene-conditioned prior knowledge into object-level representation (elaborated in Sec. \ref{sec:msa}). 
Finally the detection head regresses the bounding boxes from the location and refined features of those object candidates (elaborated in Sec. \ref{sec:detection_head}).

To enable HyperDet3D the awareness of scene-level meta information, we adopt HyperNetwork~\cite{ha2017hypernetworks} which is a neural network used to parameterize learnable parameters for another network (called primary network). 
For a target layer in primary network, its learnable parameters $\bm{W}$ are usually generated by feeding a learnable embedding $z$ or intermediate features $x$ into a hypernetwork $\bm{H}$: 
\begin{equation}
\label{eq:hypernet}
    \bm{W} = \bm{H}(z) \;\;\;\text{or} \;\;\;\bm{W} = \bm{H}(x) 
\end{equation}

Unlike conventional deep neural networks that keep the layer fixed at test time, hypernetworks enable flexibility of learnable parameters by modifying its input. 

In HyperDet3D, we propose to use a scene-conditioned hypernetwork to inject prior knowledge into the layer parameters in Transformer decoder, which dynamically adjusts the detection network in accordance with different input scenes.

\subsection{Scene-Conditioned HyperNetworks}
\label{sec:msa}
For the feature representation $\bm{o}$ of a set of object candidates produced by the backbone encoder, the goal of our scene-conditioned hypernetworks is to endow it with the prior knowledge parameterized by $\{\bm{W}, \bm{b}\}$: 
\begin{equation}
\label{eq:output_fusion_with_obj}
\hat{\bm{o}} = \bm{W}\bm{o}+\bm{b}
\end{equation}
where $\bm{W}\in \mathbb{R}^{C_{\text{out}}\times C_{\text{in}}}$ and $\bm{b}\in \mathbb{R}^{C_{\text{out}}}$ are weight and bias parameters in primary detection network. 
The parameters are produced by our scene-conditioned hypernetworks, which can be categorized into scene-agnostic and scene-specific hypernetworks. 

\textbf{Scene-Agnostic HyperNetwork: }
Take the weight parameters $\bm{W}$ of primary network for example. 
For scene-agnostic knowledge, we firstly maintain a set of $n$ scene-agnostic embedding vectors $\bm{Z}^a=\{z^a_j\in\mathbb{R}^{C_a}\}^{n}_{j=1}$. 
$\bm{Z}^a$ is then consumed by a scene-agnostic hypernetwork $\bm{h}_\theta^a$ which projects $z^a_j$ into another $\mathbb{R}^{C_{ui}}$ space, and the output $\bm{W}^a$ parameterizes our scene-agnostic knowledge:  
\begin{equation}
\label{eq:scene_agnostic_hypernetwork}
\bm{W}^a:=\{w^a_j\in \mathbb{R}^{C_{ui}}\}^{n}_{j=1},\; w^a_j = \bm{h}_\theta^a (z^a_j)
\end{equation}
where $C_{\text{ui}}$ is unit fan-in channel size, and satisfies: 
\begin{equation}
\label{eq:mod_condition}
\mod{(C_{\text{out}}, n)} \equiv 0,  \mod{(C_{\text{in}}, C_{\text{ui}})} \equiv 0
\end{equation}

While the object features are iteratively refined by a series of decoder layers~\cite{misra2021end,liu2021group}, they can be consistently incorporated with the output of scene-agnostic hypernetwork which abstracts the prior knowledge of various 3D scenes. 
In this way, we not only maintain the general scene-conditioned knowledge throughout the decoder layers, but also save the computational cost by sharing the knowledge with rich feature hierarchies. 

\textbf{Scene-Specific HyperNetwork: }
For scene-specific knowledge, we also learn a set of embedding vectors $\bm{Z}^s=\{z^s_k\in\mathbb{R}^{C_s}\}^{n}_{k=1}$ similar to $\bm{Z}^a$. 
The difference is, to adapt $\bm{Z}^s$ to the input scene, our scene-specific hypernetwork $\bm{h}_\theta^s$ uses the input scene $\bm{\text{P}}^i$ as a scene-specific query. 
Inspired by the alignment~\cite{bahdanau2015neural} in language model, we measure how well $\bm{z}_w^s$ matches the input scene (or how much they differ) in the embedding space through attention mechanism: 
\begin{equation}
\label{eq:alignment_specific}
\begin{aligned}
\bm{W}^s &:= \{w^s_k\in\mathbb{R}^{C_{ui}}\}^{n}_{k=1}\\
w_k^s &= \bm{h}_\theta^s(z_k^s, \bm{\text{P}}^i_d)
      = W_f(z_k^s||W_p\bm{\text{P}}^i_d)
\end{aligned}
\end{equation}
where $\bm{\text{P}}^i_d \in \mathbb{R}^{N_d\times 3}$, $W_p \in \mathbb{R}^{C_{n}\times N_d}$ are a subset of the current input scene, and transformation matrix which projects $\bm{\text{P}}^i_d$ into the embedding space of $\bm{Z}^s$. 
$W_f$ represents the weight matrix with Tanh the activation function. 
As we intend to get responses from the latent embedding space, we use concatenation ($\cdot||\cdot$) as coding of query points and embedding vectors similar to SDF query~\cite{park2019deepsdf}. 
We adopt the downsampled representation $\bm{\text{P}}^i_d$ instead of $\bm{\text{P}}^i$ because hypernetworks, as suggested by the previous research\cite{nirkin2021hyperseg}, do not fully capture the high-resolution information. 

From the set of scene-specific attentional scores $\bm{W}^s$ and scene-conditioned knowledge $\bm{W}^a$, now we can get unit block for $\bm{W}$: 
\begin{equation}
\label{eq:fusion_agnostic_specific}
\bm{W}^u=\bm{W}^s\odot \bm{W}^a
\end{equation}
where $\odot$ denotes the element-wise multiplication. 

\begin{figure}[tb]
\centering
\includegraphics[width=.475\textwidth]{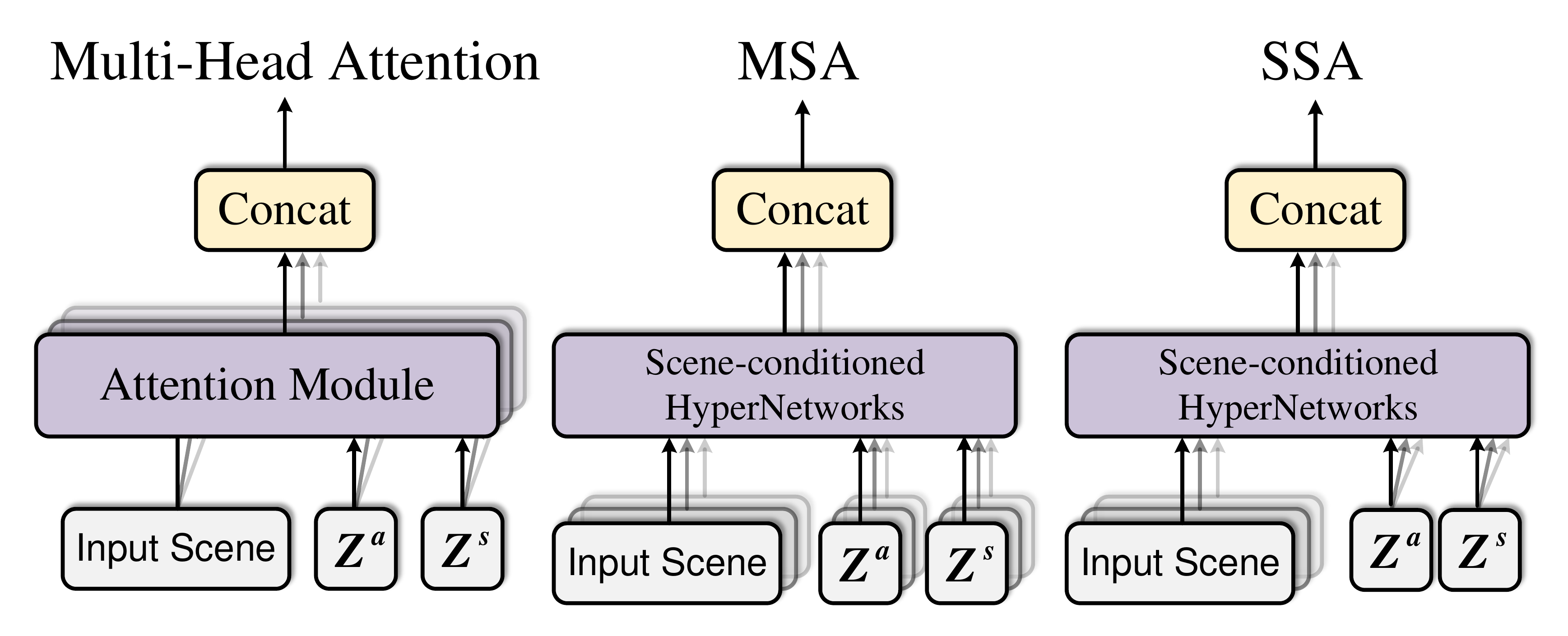}
\vspace{-7mm}
\caption{
The comparison between Multi-Head Attention~\cite{vaswani2017attention}, our proposed Multi-Head Scene-Conditioned Attention (MSA) and Single-Head Scene-Conditioned Attention (SSA). 
} \label{fig:msa}
\end{figure}

\textbf{Multi-Head Scene-Conditioned Attention: }
For the $i$-th input scene $\text{P}^i$, the abovementioned process can be encapsulated into 2 scene-conditioned attention operations: 
\begin{equation}
\label{eq:scene_conditioned_attn}
\bm{W}^u = \text{Att}_2(\{z^a_j\}, \text{Att}_1(\{z^s_k\}, \bm{\text{P}}^i))
\end{equation}
where $\text{Att}_1$ and $\text{Att}_2$ correspond to the attention in (\ref{eq:alignment_specific}) and (\ref{eq:fusion_agnostic_specific}) respectively. 
To fit the shape of target weights $\bm{W}\in\mathbb{R}^{C_{\text{out}}\times C_{\text{in}}}$ for primary network, a simple solution is to repeat $\bm{W}^u$ by $\frac{C_\text{out}}{n}\times\frac{C_\text{in}}{C_\text{ui}}$ times and tile them along its 2 dimensions. 
The feasibility is guaranteed by (\ref{eq:mod_condition}). 
As $\bm{Z}^a$ and $\bm{Z}^s$ are initialized and consumed by hypernetworks only once, we name it Single-Head Scene-Conditioned Attention (SSA). 

To allow the primary detector to jointly attend to the scene-conditioned knowledge in various sub-spaces, we further propose Multi-Head Scene-Conditioned Attention (MSA) based on SSA. 
The idea of multi-head attention is proposed in \cite{vaswani2017attention} which consumes the same set of input via parallel attention modules. 
However, as target weights $\bm{W}$ are conditioned on the input of hypernetworks in our case, we instead implement MSA by re-initializing $\bm{Z}^a$ and $\bm{Z}^s$ multiple times. 
Therefore, our MSA can be formulated as: 
\begin{equation}
\label{eq:msa}
\bm{W} = \text{Concat}(\bm{W}^u_{(1)}, \bm{W}^u_{(2)}, \dots,  \bm{W}^u_{(\frac{C_\text{out}}{n}\times\frac{C_\text{in}}{C_\text{ui}})})
\end{equation}
where $\bm{W}^u_{(l)}$ denotes the result in (\ref{eq:scene_conditioned_attn}) produced by the $l$-th initialization of $\bm{Z}^a$ and $\bm{Z}^s$. 
The Concat operation tiles the matrices along 2 dimensions similar to SSA. 

In Figure~\ref{fig:msa}, we illustrate the comparison between the original Multi-Head Attention~\cite{vaswani2017attention}, our Multi-Head Scene-Conditioned Attention (MSA) and Single-Head Scene-Conditioned Attention (SSA). 
The computation overhead for a single input sample in \cite{vaswani2017attention} is proportional to the number of parallel attention modules which define the attentional sub-spaces. 
Instead, the MSA network is shared between all training samples in our HyperDet3D. 
Moreover, as we mine the sub-spaces via hypernetwork structures, MSA exploits the flexibility of scene-conditioned knowledge via modifying the input in (\ref{eq:hypernet}). 
In comparison, SSA consumes the same set of embedding vectors and is inferior to MSA in terms of expressiveness, which we verify in the ablation experiments.

The pipeline of obtaining the bias parameters $\bm{b}$ is similar to that of $\bm{W}$, which we display in the supplementary pages. 
$\bm{W}$ and $\bm{b}$ are aggregated with object features as in (\ref{eq:output_fusion_with_obj}). 
The renewed representation $\hat{\bm{o}}$ is then consumed by the detection head to generate the detection results.

\subsection{Disentangled Detection Head}
\label{sec:detection_head}
Following \cite{qi2019deep}, existing works locate the object center $\bm{c}_i$ via directly regressing an offset ($\Delta\bm{q}_i$) from the candidate location $\bm{q}_i$ by a detection head parameterized by $\bm{W}_c$: 
\begin{equation}
\label{eq:vote_offset}
\bm{c}_i = \bm{q}_i + \Delta \bm{q}_i, \;\Delta \bm{q}_i=\bm{W}_c \hat{\bm{o}}_i
\end{equation}
Here we use a Disentangled variant of Detection Head (DDH) which factorizes the offset regression into 2 branches. 
Given a predicted $\Delta\bm{q}_i$, one branch regresses a scalar $r\in \mathbb{R}^1$ to modulate its length, and another regresses a 4-dim vector, $\bm{R} \in \mathbb{R}^4$, to modulate its orientation. 
Each branch contains a light-weighted regression head. 
$\bm{R}$ is regarded as the real part of a quaternion, which can be transformed into a rotation matrix to modulate the orientation of $\Delta\bm{q}_i$. 
Therefore, the final offset $\Delta\bm{q}_i^{'}$  is computed as follows: 
\begin{equation}
\Delta\bm{q}_i^{'} = f_T(\bm{R})*(r\Delta\bm{q}_i)
\end{equation}
where $*$ denotes dot production. 
$f_T$ is the transformation function defined in \cite{shoemake1985animating} which converts the quaternion into a 3x3 rotation matrix. 
Note that $\bm{R}$ is firstly L2-normalized when being transformed.

\subsection{Implementation Details}
The backbone network PointNet++~\cite{qi2017pointnet++} in HyperDet3D contains 4 set abstraction layers which downsample the input scan into \{2048,1024,512,256\} points consecutively. 
The radius for ball query is \{0.2m,0.4m,0.8m,1.2m\}. 
Then 2 feature propagation layers recover them into 1024 points and produce the point-wise features. 
We use the KPS proposed in \cite{liu2021group} to generate object candidates from the original locations of these 1024 points, as it saves the computational cost in O$(N^2)$ search space of FPS~\cite{qi2017pointnet++}. 

To obtain $\bm{o}$ in each decoder layer, we follow \cite{misra2021end,liu2021group} to employ the standard multi-head attention layer to compute the self-attention of object candidates, followed by the cross-attention between object candidates and downsampled points produced by the backbone. 
The scene-agnostic hypernetwork in (\ref{eq:scene_agnostic_hypernetwork}) contains 2 linear layers. 
The scene-specific hypernetwork in (\ref{eq:alignment_specific}) contains 1 linear layer followed by Tanh activation function. 
Each linear layer is parameterized by a weight matrix and bias vector, initialized by Xavier~\cite{glorot2010understanding} and zeros. 
For the scene query $\bm{\text{P}}^i_d$ of scene-specific hypernetwork, we use the off-the-shelf downsampled results of KPS. 

As for detection head, each light-weighted regression head mainly contains a fully-connected (FC) layer to map $\bm{f}_i$ into $r$ or $\bm{R}$. 
In $r$-head, the output of FC layer is processed by sigmoid function and further normalized into [0.9,1.1] to control the extent of adjustment. 
In $\bm{R}$-head, identity quaternion is added to $\bm{R}$ before transformation ($f_T$), which can simultaneously hold the possibility of identity rotation and control the rotation degree.

\section{Experiment}
In experiment section, we firstly introduce the datasets and evaluation metrics of the benchmark for 3D object detection (Sec.~\ref{sec:datasets_protocols}). 
We then display the thorough experimental results by comparing HyperDet3D with state-of-the-art approaches both quantitatively and qualitatively (Sec.~\ref{sec:main_results}). 
We also analyze the design choice and effectiveness of HyperDet3D by ablation studies and cross-dataset evaluation (Sec.~\ref{sec:ablation_discussion}). 
Finally we point out the limitation of our work (Sec.~\ref{sec:limitation}). 
More analysis and visualizations are provided in the supplementary pages. 

\subsection{Datasets and Settings}
\label{sec:datasets_protocols}
\textbf{ScanNet V2: }
The ScanNet V2 dataset~\cite{dai2017scannet} includes 1,513 scanned and reconstructed indoor scenes, with axis-aligned bounding box labels for 18 object
categories. 
The point cloud data are converted from reconstructed meshes. 
Following \cite{qi2019deep}, we employ 1,201 scenes as the training set and the rest 312 validation scenes as the test set. 

\textbf{SUN RGB-D V1: }
The SUN RGB-D V1 dataset~\cite{song2015sun} contains 10k single-view indoor RGB-D images, 5,285 for training and 5,050 for testing. 
It's densely annotated with 64k oriented 3D bounding boxes. 
The whole dataset is categorized into 37 indoor object classes. 
For fair comparison, we follow the evaluation protocol in \cite{qi2019deep} which selects the 10 most common categories. 

For both datasets, we only employ point cloud data as the input. 
No scene-level supervision is employed by HyperDet3D. 
Following \cite{qi2019deep}, we report the detection performance on the validation sets by computing mean Average Precision (mAP) with 3D IoU threshold 0.25 (mAP@0.25) and 0.5 (mAP@0.5). 
Detection performance on individual categories and their average results are displayed. 

As for the training strategy, in the first 100 epochs of both datasets, the detection head directly consumes $\bm{o}$ rather than $\hat{\bm{o}}$ in (\ref{eq:output_fusion_with_obj}). 
Then the network was finetuned for 300 and 500 epochs on ScanNet and SUN RGB-D respectively, using $\hat{\bm{o}}$ instead. 
The strategy aims for the stability of loss curves when incorporated with hypernetworks. 
The finetuned network was used for inference at test time. 
The details of hyper-parameters for 2 datasets can be found in the supplementary material. 

\subsection{Main Results}
\label{sec:main_results}
\textbf{Quantitative results: }
We compare our HyperDet3D quantitatively with a number of reference methods, which can be divided into 3 categories: early approaches that require 2D guidance to locate 3D objects~\cite{song2016deep,qi2018frustum,yi2019gspn,hou20193d,ren2016three,lahoud20172d,xu2018pointfusion}, voting-based approaches that explore optimal local representation to provide informative cues~\cite{qi2019deep,chen2020hierarchical,zhang2020h3dnet,cheng2021back,xie2021venet,najibi2020dops,liu2020group,qi2020imvotenet}, and relation-based approaches that explore the interaction between local elements such as objects or point clusters~\cite{xie2020mlcvnet,du2020spot,pan20213d,misra2021end,liu2021group,feng2020relation}. 
The experimental results are shown in Table~\ref{tab:sota-table} and Table~\ref{tab:ScanNetCat}. 
Bold indicates the best results under the corresponding metrics. 

From the comparison results in Table~\ref{tab:sota-table}, we can observe that the state-of-the-art relation-based GF3D~\cite{liu2021group} outperforms all the other compared methods, except for ImvoteNet~\cite{qi2020imvotenet} which incorporates 2D image votes. 
However, thanks to the acquired scene-conditioned prior knowledge, our HyperDet3D stills achieves leading average on 2 metrics of both ScanNet V2 (+1.8\% mAP@0.25, +4.4\% mAP@0.5) and SUN RGB-D V1 (+0.5\% mAP@0.25, +2.1\% mAP@0.5) validation set. 
Note that compared with SUN RGB-D, ScanNet is annotated with 1.8x as many categories for 3D detection task. 
Therefore, the scene-level prior knowledge learned by HyperDet3D content is relatively richer in ScanNet than SUN RGB-D, and yields more significant mAP gain on the former dataset. 

\begin{table*}[t]
\small
\caption{
3D object detection results on the ScanNet V2 validation set (left) and the SUN RGB-D V1 validation set (right).
Evaluation metric is average precision with 3D IoU thresholds as 0.25 and 0.50.
Results of H3DNet~\cite{zhang2020h3dnet} are reported under $4$ PointNet++ backbones settings.
Results of 3DETR~\cite{misra2021end} are reported on its stronger 3DETR-m variant with inductive biases. 
}
\label{tab:sota-table}
\vspace{-3mm}
\hspace{2mm}
\begin{minipage}[t]{0.47\linewidth}
\centering
\resizebox{\textwidth}{38mm}{
\begin{tabular}{l|c|c|c}
\hline
 ScanNet V2 & Input & mAP@0.25 & mAP@0.50 \\ \hline
DSS~\cite{song2016deep}  & Geo + RGB  & 15.2 & 6.8 \\
MRCNN~\cite{he2017mask} & Geo + RGB & 17.3 & 10.5 \\
F-PointNet~\cite{qi2018frustum}  & Geo + RGB  & 19.8 & 10.8 \\
GSPN~\cite{yi2019gspn}  & Geo + RGB  & 30.6 & 17.7 \\
3D-SIS~\cite{hou20193d}  & Geo + 5 views  & 40.2 & 22.5 \\
\hline
VoteNet~\cite{qi2019deep} & Geo only & 58.6 & 33.5 \\
GCENet~\cite{liu2020group} & Geo only & 60.7 & - \\
HGNet~\cite{chen2020hierarchical} & Geo only & 61.3 & 34.4 \\
DOPS~\cite{najibi2020dops} & Geo only & 63.7 & 38.2 \\
H3DNet*~\cite{zhang2020h3dnet} & Geo only & 67.2 & 48.1 \\
BRNet~\cite{cheng2021back} & Geo only & 66.1 & 50.9 \\
VENet~\cite{xie2021venet} & Geo only & 67.7 & - \\
\hline
RGNet~\cite{feng2020relation} & Geo only & 48.5 & 26.0 \\
SPOT~\cite{du2020spot} & Geo only & 59.8 & 40.4 \\
MLCVNet~\cite{xie2020mlcvnet} & Geo only & 64.7 & 42.1 \\
PointFormer~\cite{pan20213d} & Geo only & 64.1 & 42.6 \\
3DETR*~\cite{misra2021end} & Geo only & 65.0 & 47.0 \\
GF3D~\cite{liu2021group} & Geo only & 69.1 & 52.8 \\
\hline
Ours & Geo only & \textbf{70.9} & \textbf{57.2} \\
\hline
\end{tabular}}
\end{minipage}
\hspace{4mm}
\begin{minipage}[t]{0.47\linewidth}
\centering
\resizebox{\textwidth}{38mm}{
\begin{tabular}{l|c|c|c}
\hline
SUN RGB-D & Input & mAP@0.25 & mAP@0.50 \\ \hline
DSS~\cite{song2016deep} & Geo + RGB & 42.1 & - \\
2D-driven~\cite{lahoud20172d} & Geo + RGB & 45.1 & - \\
PointFusion~\cite{xu2018pointfusion} & Geo + RGB & 45.4 & - \\
COG~\cite{ren2016three} & Geo + RGB & 47.6 & - \\
F-PointNet~\cite{qi2018frustum} & Geo + RGB & 54.0 & - \\
\hline
VoteNet~\cite{qi2019deep} & Geo only & 57.7 & 32.9 \\
H3DNet*~\cite{zhang2020h3dnet} & Geo only & 60.1 & 39.0 \\
VENet~\cite{xie2021venet} & Geo only & 62.5 & 39.2 \\
GCENet~\cite{liu2020group} & Geo only & 60.8 & 40.1 \\
HGNet~\cite{chen2020hierarchical} & Geo only & 61.6 & - \\
ImVoteNet~\cite{qi2020imvotenet} & Geo + RGB & 63.4 & - \\
BRNet~\cite{cheng2021back} & Geo only & 61.1 & 43.7 \\
\hline
3DETR*~\cite{misra2021end} & Geo only & 59.1 & 32.7 \\
RGNet~\cite{feng2020relation} & Geo only & 59.2 & - \\
MLCVNet~\cite{xie2020mlcvnet} & Geo only & 59.8 & - \\
SPOT~\cite{du2020spot} & Geo only & 60.4 & 36.3 \\
PointFormer~\cite{pan20213d} & Geo only & 61.1 & 36.6 \\
GF3D~\cite{liu2021group} & Geo only & 63.0 & 45.2 \\
\hline 
Ours & Geo only & \textbf{63.5} & \textbf{47.3} \\
\hline
\end{tabular}}
\end{minipage}
\end{table*}

We then look into the per-category results of mAP@0.5 on ScanNet V2 validation set, which is the benchmark with more categories, more challenging threshold for evaluation, and more performance gain by our method. 
The detailed results are displayed in Table~\ref{tab:ScanNetCat}. 
For the categories largely conditioned on the scene prior (such as \textit{bed} in bedroom, \textit{fridge} in kitchen/canteen, \textit{shower curtain}/\textit{toilet}/\textit{sink}/\textit{bathtub} in bathroom), they consistently obtain notable AP gain compared with the baseline methods. 
This indicates the effectiveness of learned scene-conditioned knowledge by HyperDet3D. 
The performance drops on the \textit{counter} category which is less conditioned on the scene-level semantics. 
We display the detailed results on SUN RGB-D in the supplementary pages. 

In Table~\ref{tab:params_num}, we compare our method with the state-of-the-art GF3D~\cite{liu2021group} furtherly\footnote{In Table~\ref{tab:params_num}, as suggested by \cite{liu2021group}, L denotes the number of decoders; O denotes the number of object candidates; and w2$\times$ denotes the feature dimension in backbone is expanded by 2 times. }. 
It can be seen that in a normal or light-weighted version of network configurations, our approach outperforms GF3D in both metrics while containing notably fewer learnable parameters. 
Therefore, HyperDet3D is likely to efficiently absorb the external data due to the mechanism of scene-conditioned hypernetworks and knowledge sharing in different layers. 

\begin{table*}[t!]
\caption{3D object detection results on the ScanNet V2 validation dataset. We show per-category results of mean average precision (mAP) with 3D IoU threshold 0.5 as proposed in \cite{song2015sun}, and mean of AP across all semantic classes with 3D IoU threshold 0.5.} 
\vspace{-3mm}
\begin{adjustbox}{width=.98\textwidth,center}
 \begin{tabular}{l | c  c  c  c  c  c  c  c  c  c  c  c  c  c  c  c  c  c | c } 
 \hline
 & cab & bed & chair & sofa & tabl & door & wind & bkshf & pic & cntr & desk & curt & fridg & showr & toil & sink & bath & ofurn & mAP\\
 \hline
Votenet~\cite{qi2019deep} & 8.1 & 76.1 & 67.2 & 68.8 & 42.4 & 15.3 & 6.4 & 28.0 & 1.3 & 9.5 & 37.5 & 11.6 & 27.8 & 10.0 & 86.5 & 16.8 & 78.9 & 11.7 & 33.5 \\
DOPS~\cite{najibi2020dops} & 25.2 & 70.2 & 75.8 & 54.8 & 41.2 & 27.8 & 12.1 & 21.4 & \textbf{12.3} & 9.5 & 39.4 & 24.4 & 33.7 & 17.3 & 80.6 & 35.7 & 71.0 & 35.0 & 38.2\\
MLCVNet~\cite{xie2020mlcvnet} & 16.6 & 83.3 & 78.1 & 74.7 & 55.1 & 28.1 & 17.0 & 51.7 & 3.7 & 13.9 & 47.7 & 28.6 & 36.3 & 13.4 & 70.9 & 25.6 & 85.7 & 27.5 & 42.1 \\
PointFormer~\cite{pan20213d} & 19.0 & 80.0 & 75.3 & 69.0 & 50.5 & 24.3 & 15.0 & 41.9 & 1.5 & 26.9 & 45.1 & 30.3 & 41.9 & 25.3 & 75.9 & 35.5 & 82.9 & 26.0 & 42.6 \\
H3DNet~\cite{zhang2020h3dnet} & 20.5 & 79.7 & 80.1 & 79.6 & 56.2 & 29.0 & 21.3 & 45.5 & 4.2 & 33.5 & 50.6 & 37.3 & 41.4 & 37.0 & 89.1 & 35.1 & 90.2 & 35.4 & 48.1 \\
BRNet~\cite{cheng2021back} & 28.7 & 80.6 & 81.9 & 80.6 & 60.8 & 35.5 & 22.2 & 48.0 & 7.5 & \textbf{43.7} & 54.8 & 39.1 & 51.8 & 35.9 & 88.9 & 38.7 & 84.4 & 33.0 & 50.9 \\
GF3D~\cite{liu2021group} & 26.0 & 81.3 & 82.9 & 70.7 & \textbf{62.2} & 41.7 & 26.5 & \textbf{55.8} & 7.8 & 34.7 & 67.2 & 43.9 & 44.3 & 44.1 & 92.8 & 37.4 & 89.7 & 40.6 & 52.8 \\
\hline
Ours & \textbf{33.1} & \textbf{90.1} & \textbf{83.8} & \textbf{83.8} & 60.3 & \textbf{43.6} & \textbf{31.7} & 52.2 & 4.2 & 20.9 & \textbf{78.5} & \textbf{49.0} & \textbf{61.1} & \textbf{56.3} & \textbf{95.9} & \textbf{43.9} & \textbf{100} & \textbf{42.3} & \textbf{57.3} \\
\hline
\end{tabular}
\label{tab:ScanNetCat}
\end{adjustbox}
\end{table*}

\begin{table}[t]
    \centering
    \caption{Comparison with GroupFree-3D~\cite{liu2021group} (GF3D) with various configurations on the ScanNet V2 validation set.
    The upper section shows results for GF3D models reported in \cite{liu2021group}. 
    }
    \vspace{-3mm}
    \begin{adjustbox}{width=.42\textwidth,center}
    \begin{tabular}{rccHc}
        \toprule
        Model & backbone & \#params & mAP@0.25 & mAP@0.5 \\
        \midrule
        GF3D-(L6,O256) & PointNet++ & 14.5M & 67.3 & 48.9 \\
        GF3D-(L12,O512) & PointNet++w2$\times$ & 29.6M & 69.1 & 52.8 \\
        \midrule
        Ours-(L6,O256) & PointNet++ & 11.1M &   &  51.0\\
        Ours-(L12,O512) & PointNet++w2$\times$ & 22.6M & \textbf{70.9}  & 57.2 \\
        \bottomrule
    \end{tabular}
    \end{adjustbox}
    \label{tab:params_num}
\end{table}

\begin{figure*}[tb]
\centering
\includegraphics[width=0.99\textwidth]{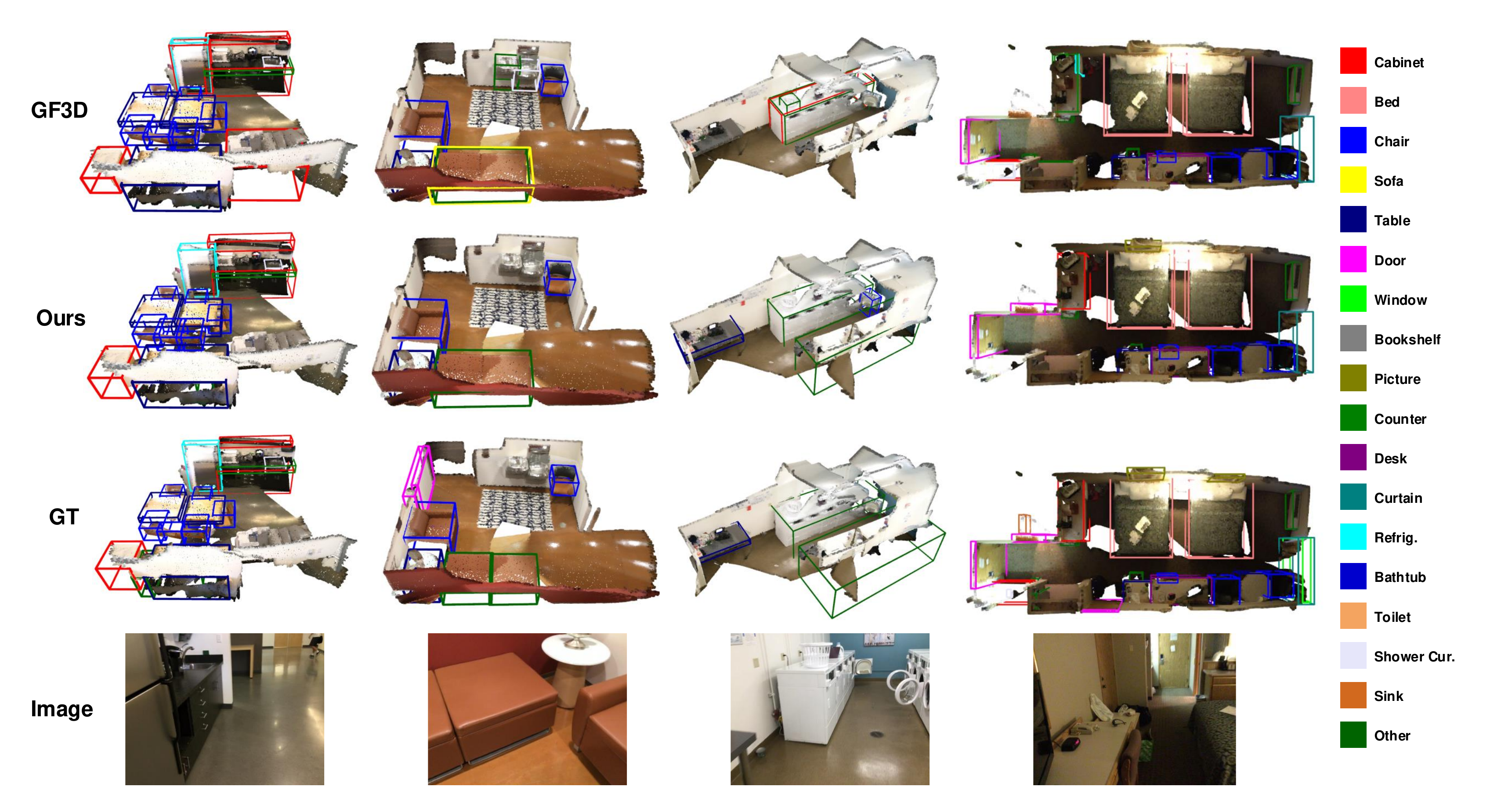}
\vspace{-4mm}
\caption{
Qualitative comparisons between our approach and GF3D~\cite{liu2021group} baseline approach on the ScanNet V2 validation set. 
The groundtruth annotations (GT) and 2D image scans are taken as reference. 
Our method achieves favourable results compared to the baseline method. 
Fewer ambiguous detections are observed in our results. 
The point clouds are colored only for easy illustration, and not utilized in the compared method nor ours. 
(\textit{Best viewed in color.})
} \label{fig:qualitative}
\end{figure*}

\textbf{Qualitative Results: }
In Figure~\ref{fig:qualitative}, we illustrate the representative 3D object detection results of 4 scans in ScanNet V2 validation set. 
Taking groundtruth annotations (GT) and real image scans as reference, we compare our HyperDet3D with the state-of-the-art GF3D~\cite{liu2021group} which involves the dense interaction between object candidates. 
The first 3 scans highlight the ambiguity in the aspect of largely intersected bounding boxes, where the baseline module mistakes a refrigerator or washing machine for a cabinet, or detect a sink in an office. 
With the help of scene-conditioned prior knowledge, our HyperDet3D can obtain better detection results on these objects. 
The ambiguous detections also include the mistaken detections. 
For example, in the last scan, the baseline method mistakes a cabinet in a bedroom for a counter. 

\subsection{Ablation Study and Discussions}
\label{sec:ablation_discussion}
To analyze the importance of learned scene-conditioned knowledge in our HyperDet3D network, we conducted ablation experiments on various combinations of design choices. 
The quantitative results are shown in Table~\ref{tab:ablation}. 
The baseline model only uses disentangled detection head and we gray its corresponding row for clear comparison. 
Applying the SSA to learn scene-conditioned knowledge leads to improvement of mAP@0.25 by 1.2\%,  and mAP@0.5 by 1.2\%. 
The multi-head variant (MSA) further brings +1.1\% mAP@0.25 and +2.5\% mAP@0.5. 
As expected, only learning scene-agnostic or scene-specific prior knowledge is inadequate for thorough scene-conditioned understanding. 
For the challenging mAP@0.5 metric, only learning scene-agnostic or scene-specific knowledge causes performance drop by -1.8\% and -3.4\% respectively. 
The removal of disentangled regression of center offsets leads to a performance drop by -0.6\% mAP@0.25 and -0.4\% mAP@0.5, which indicates that the exquisite regression of targets helps to leverage the learned scene-conditioned knowledge.

\begin{table}[t]
\centering
\caption{
Experimental results of ablation studies on the ScanNet V2 validation set. 
The baseline method only applies the disentangled detection head (DDH) on candidate features without scene-conditioned prior knowledge (the row colored in gray). 
}
\vspace{-3mm}
\begin{adjustbox}{width=.46\textwidth,center}
\begin{tabular}{cc|cc|c|cc}
\toprule
\multicolumn{2}{c|}{Scene-Conditioned} & \multicolumn{2}{c|}{Attention} & \multirow{2}*{DDH} & \multirow{2}*{mAP@0.25} & \multirow{2}*{mAP@0.5} \\ \cline{1-4} 
agnostic   & specific   &     SSA    &     MSA    &            &                               \\ \hline
\rowcolor{gray!30}
           &            &            &            & \checkmark &     68.6      &      53.5     \\
\checkmark & \checkmark & \checkmark &            & \checkmark &     69.8      &      54.7     \\
           & \checkmark &            & \checkmark & \checkmark &     70.6      &      55.4     \\
\checkmark &            &            & \checkmark & \checkmark &     70.0      &      53.8     \\
\checkmark & \checkmark &            & \checkmark &            &     70.3      &      56.8     \\
\checkmark & \checkmark &            & \checkmark & \checkmark & \textbf{70.9} & \textbf{57.2} \\
\bottomrule
\end{tabular}
\end{adjustbox}
\label{tab:ablation}
\end{table}
\begin{table}[t!]
\caption{
Cross-dataset evaluation results on the ScanNet V2 val dataset, which is pre-trained on the SUN RGB-D V1 val dataset. 
We show mAPs of 8 shared categories between ScanNet V2 and SUN RGB-D, and all 18 categories of ScanNet V2. The 3D IoU threshold of mAP is 0.5. 
}
\vspace{-3mm}
\begin{adjustbox}{width=.47\textwidth,center}
 \begin{tabular}{l | c  c  c  c  c  c  c  c | c c} 
 \hline
 & bed & chair & sofa & tabl & bkshf & desk & toil & bath &  mAP$_{8}$ & mAP$_{18}$\\
 \hline
VoteNet & 30.3 & 21.7 & 12.4 & 8.3 & \textbf{4.4} & 4.4 & 21.7 & 33.4 & 17.1& 8.4\\
GF3D &  66.6 & 21.3 & 46.9 & \textbf{17.8} & 0.4 & 25.6 & 54.6 & 48.6 & 35.2 &19.1\\
\hline
Ours  & \textbf{78.9} & \textbf{22.7} & \textbf{58.0} & 16.0 & 2.4 & \textbf{40.1} & \textbf{58.9} & \textbf{71.4} & \textbf{43.6} &\textbf{22.2}\\
\hline
\end{tabular}
\label{Result:Cross_dataset}
\end{adjustbox}
\end{table}

\textbf{Cross-dataset evaluation: }
Since HyperDet3D learns the scene-conditioned knowledge as prior, we infer such knowledge acquired by the detector still takes effect when faced with domain gaps. 
To validate this, we conducted cross-dataset evaluation in comparison with VoteNet~\cite{qi2019deep} and GF3D~\cite{liu2021group} as the baseline detectors. 
We firstly pre-trained the baseline detectors and HyperDet3D on the SUN RGB-D V1 validation set then finetuned on the ScanNet V2 validation set. 
The backbone networks in all 3 approaches and the scene-conditioned hypernetworks in ours were frozen during finetuning. 

In Table~\ref{Result:Cross_dataset}, we show the detection mAP of 8 shared categories between SUN RGB-D and ScanNet with IoU threshold set as 0.5, as well as average mAP over all 18 (mAP$_8$) categories in ScanNet or the shared 8 (mAP$_{18}$) categories. 
The observation is two-fold. 
Our HyperDet3D surpasses the baseline methods on both mAP$_8$ and mAP$_{18}$, especially on the shared categories between 2 datasets. 
This indicates that the scene-conditioned knowledge learned on the source dataset can be well transferred to the target dataset by our method. 
On the other hand, among the 8 shared categories, those more conditioned on the scene semantics are improved by an obvious margin similar to the results in Table~\ref{tab:sota-table}. 
The exception is the \textit{bookshelf} category partially due to the scarcity of library scenes (1.9\%) in SUN-RGBD~\cite{song2015sun}. 
Moreover, the novel categories such as \textit{refrigerator} and \textit{sink} are improved by +14.9\% and +11.1\% respectively. 
The details can be found in supplementary pages. 

\textbf{Incorporation of scene labels: }
An interesting question is, what if we utilize the groundtruth scene label as an additional supervision? To this end, we added a classification(Cls.) branch to the bottleneck of the backbone in GF3D, and finetuned the whole network for 100 epochs based on the GF3D pretrained model. 
The ScanNet results in Table~\ref{tab:cls_baseline} suggest the additional branch with scene type labels improves the detection performance, but is still inferior to HyperDet3D. 
Note that our HyperDet3D achieves the best results without any supervision on the scene type classification. 
We expect better detection performance by training a unique detector for each type of scene. 
However, this may limit the generality of the method and is less computationally friendly for real-world applications. 

\begin{table}[t!]
\centering
\caption{
Comparison with Multi-task classification (Cls.) baseline. 
}
\vspace{-3mm}
\begin{adjustbox}{width=.46\textwidth,center}
\begin{tabular}{cc|cc|cc}
\toprule
\multicolumn{2}{c|}{HyperDet3D} & \multicolumn{2}{c|}{GF3D} & \multicolumn{2}{c}{GF3D+Cls.} \\
mAP@0.25 & mAP@0.5 & mAP@0.25 & mAP@0.5 & mAP@0.25 & mAP@0.5 \\ \hline
\textbf{70.9} & \textbf{57.2} & 69.1 & 52.8 & 69.4 & 54.3 \\
\bottomrule
\end{tabular}
\end{adjustbox}
\label{tab:cls_baseline}
\end{table}

\subsection{Limitations}
\label{sec:limitation}
As we focus on the scene-level information, we can observe some failure cases on detailed local geometries. 
For example, in the second example of Figure~\ref{fig:qualitative}, HyperDet3D misdetects 2 closely connected objects as a whole. 
A possible solution is to incorporate more detailed representation of scene query, which may require SDF~\cite{park2019deepsdf} to exquisitely model the geometries in the scene. 
Moreover, another important work Mix3D~\cite{nekrasov2021mix3d} proposes to reduce the scene-level variation by enriching the scene-level data with object-level information, while HyperDet3D aims to utilize such scene-specific variation by endowing object representation with scene-prior knowledge. 
We expect a future solution might combine the advantages of both methods.

\section{Conclusion}
In this paper we have introduced HyperDet3D: a new framework to explore scene-conditioned prior for 3D object detection. 
Our HyperDet3D simultaneously learns the scene-agnostic knowledge which explores the sharable abstracts from various 3D scenes, and scene-specific knowledge which adapts the detector to the given scene. 
HyperDet3D achieves state-of-the-art results on the 3D object detection benchmark of 2 widely-used datasets, and demonstrates effectiveness when faced with domain gap. 

\textbf{Potential Impact:} Our method aims to improve the researches on 3D object detection, which is critical for the safety of robotic systems. 
Similar to many deep learning methods, one potential negative impact is that it still lacks theoretical guarantees. 
To improve the applicability in this domain, the community might consider challenges of explainability and transparency. 

\textbf{Acknowledgements: }This work was supported in part by the National Key Research and Development Program of China
under Grant 2017YFA0700802, in part by the National Natural Science Foundation of China under
Grant 62125603, Grant U1813218, in part by Beijing Academy
of Artificial Intelligence (BAAI), and in part by a grant from the Institute for Guo Qiang, Tsinghua
University.

\clearpage

\begin{table*}[bt!]
\caption{3D object detection results on SUN RGB-D V1 validation dataset. We show per-category results of mean average precision (mAP) with 3D IoU threshold 0.5 as proposed in \cite{song2015sun}, and mean of AP across all semantic classes with 3D IoU threshold 0.5.} 
\vspace{-3mm}
\begin{adjustbox}{width=0.75\textwidth,center}
 \begin{tabular}{l | c  c  c  c  c  c  c  c  c  c | c } 
 \hline
 & bathtub & bed & bookshelf & chair & desk & drser & nightstand & sofa & table & toilet & mAP\\
 \hline
Votenet~\cite{qi2019deep} & 45.4 & 53.4 & 6.8 & 56.5 & 5.9 & 12.0 & 38.6 & 49.1 & 21.3 & 68.5 & 35.8\\
H3DNet~\cite{zhang2020h3dnet} & 47.6 & 52.9 & 8.6 & 60.1 & 8.4 & 20.6 & 45.6 & 50.4 & 27.1 & 69.1 & 39.0 \\
BRNet~\cite{cheng2021back} & 55.5 & 63.8 & 9.3 & 61.6 & 10.0 & \textbf{27.3} & 53.2 & 56.7 & 28.6 & 70.9 & 43.7 \\
GF3D~\cite{liu2021group} & \textbf{64.0} & 67.1 & 12.4 & \textbf{62.6} & \textbf{14.5} & 21.9 & 49.8 & 58.2 & \textbf{29.2} & 72.2 & 45.2 \\
\hline
Ours & 59.1 & \textbf{69.4} & \textbf{21.1} & 62.1 & 11.3 & 14.0 & \textbf{57.4} & \textbf{61.3} & 27.6 & \textbf{89.8} & \textbf{47.3} \\
\hline
\end{tabular}
\label{tab:SunRGBDCat}
\end{adjustbox}
\end{table*}

\begin{table*}[bt!]
\caption{3D object detection results on the ScanNet V2 validation dataset. We show per-category results of mean average precision (mAP) with 3D IoU threshold 0.5 as proposed in \cite{song2015sun}, and mean of AP across all semantic classes with 3D IoU threshold 0.5.} 
\vspace{-4mm}
\begin{adjustbox}{width=.98\textwidth,center}
 \begin{tabular}{l | c  c  c  c  c  c  c  c  c  c  c  c  c  c  c  c  c  c | c } 
 \hline
 & cab & bed & chair & sofa & tabl & door & wind & bkshf & pic & cntr & desk & curt & fridg & showr & toil & sink & bath & ofurn & mAP\\
 \hline
GF3D~\cite{liu2021group} & 5.3 & 66.6 & 21.3 & 46.9 & 17.8 & 3.3 & 4.5 & 0.4 & 0 & 8.0 & 25.6 & 6.7 & 22.3 & 0 & 54.6 & 11.7 & 48.6 & 0.4 & 19.1 \\
Ours & 10.8 & 78.9 & 22.7 & 58.0 & 16.0 & 2.9 & 2.2 & 2.4 & 0.5 & 13.3 & 40.1 & 11.6 & 7.4 & 0 & 58.9 & 0.5 & 71.4 & 2.3 & 22.2 \\
\hline
\end{tabular}
\label{tab:ScanNetCat_cross}
\end{adjustbox}
\end{table*}
\begin{appendix}
\section{Technical Details}
Here we introduce some techinical details on the proposed HyperDet3D, including the processing of obtaining $\bm{b}$, and some hyper-parameters for the experiments on 2 datasets. 
\subsection{Processing of obtaining biases}
For bias $\bm{b}$, we maintain $\bm{Z}^a$ and $\bm{Z}^s$ which are the same as in obtaining $\bm{W}$. 
The scene-specific and scene-agnostic hypernetworks both contain a single linear layer. 
Before the second attention, we take the average results of scene-specific and scene-agnostic vectors along the feature dimension. 

\subsection{Hyper-parameters}
Following \cite{liu2021group}, the number of decoder layers for ScanNet~\cite{dai2017scannet} and SUN RGB-D~\cite{song2015sun} are 12 and 6 respectively. 
We set $C_a$=256 and $n$=256 for both datasets. 
We set $C_s$=253 and 285 for ScanNet and SUN RGB-D respectively. 

\section{Other Experimental Results}
\subsection{Detailed Results on SUN RGB-D}

As shown in Table~\ref{tab:SunRGBDCat}, we detail the per-category results on the SUN RGB-D~\cite{song2015sun} dataset. 
We observe the obvious improvements on category \textit{bed}, \textit{bookshelf}, \textit{nightstand} and \textit{toilet} which are more conditioned on its corresponding scenes (beddroom, washroom, etc.). 
The exception is the \textit{bathtub} category with performance drop of -4.9\%, which might be attributed to the extremely scarcity of its annotations in the dataset (the least one in the statistical distribution shown in the paper~\cite{song2015sun}). 

\subsection{Detailed Results on Cross-dataset Evaluation}
For the cross-dataset experiments where we pretrained HyperDet3D on SUN RGB-D and finetuned on ScanNet v2, we display the detailed per-category results in Table~\ref{tab:ScanNetCat_cross}, including the extra 10 novel categories. 

\subsection{Ablation Study on Cross-dataset Evaluation}
The removal of scene-specific knowledge and MSA$\rightarrow$SSA brings -3.0\% and -3.4\% on mAP$_8$ respectively. 
We infer the expressiveness of MSA contributes more than scene-specific knowledge when tackling domain gap. 

\section{More Visualization Results}
\begin{figure*}[tb]
\centering
\includegraphics[width=0.95\textwidth]{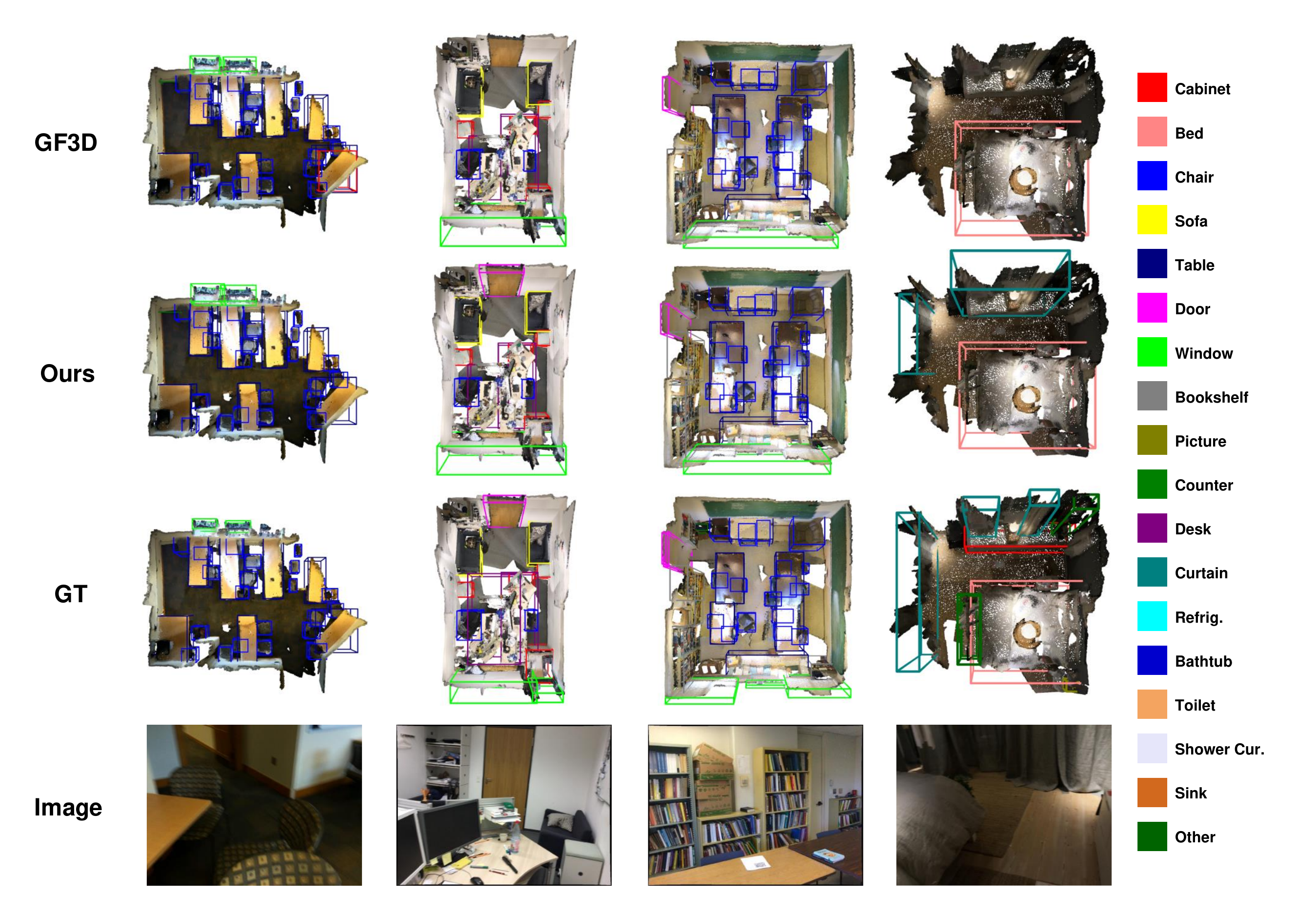}
\vspace{-2mm}
\caption{
Qualitative detection results on the ScanNet V2 validation set. 
(\textit{Best viewed in color.})
} \label{fig:qualitative_scannet_supp}
\end{figure*}

\begin{figure*}[tb]
\centering
\includegraphics[width=0.95\textwidth]{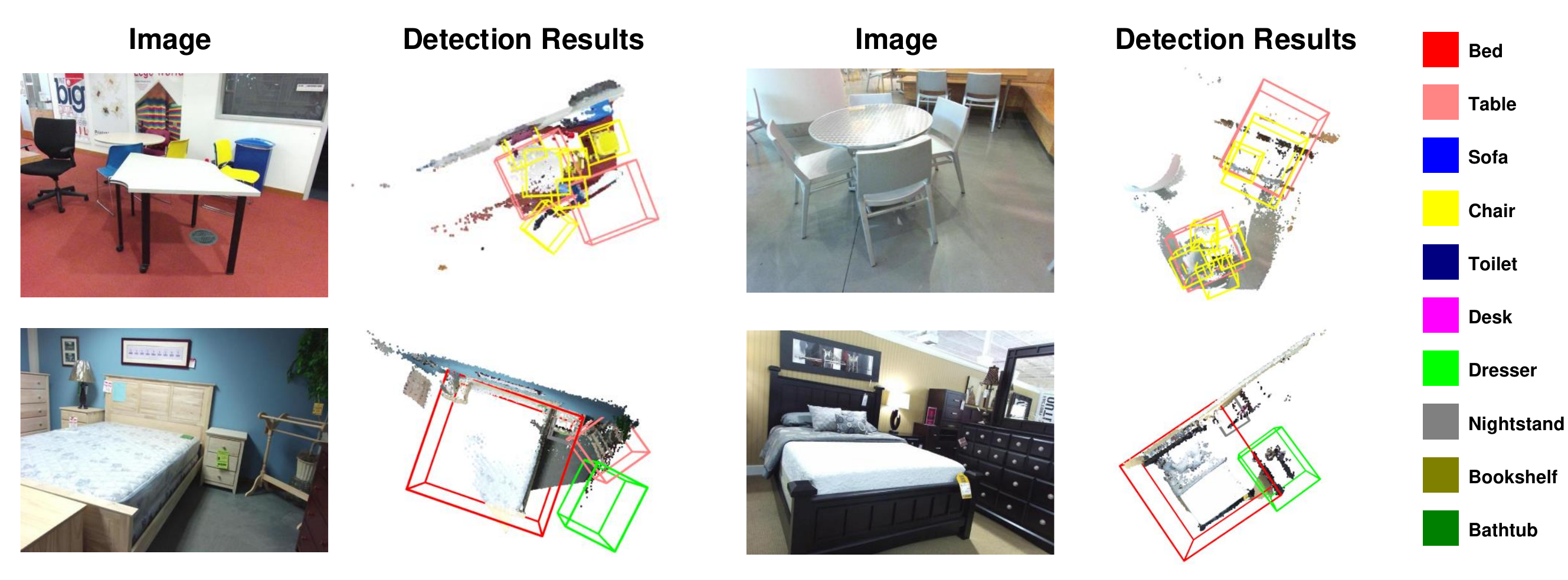}
\vspace{-2mm}
\caption{
Qualitative detection results on the SUN RGB-D V1 validation set. 
(\textit{Best viewed in color.})
} \label{fig:qualitative_sunrgbd_supp}
\end{figure*}

We have additionally illustrated qualitative results on ScanNet v2 and SUN RGB-D. 
The results on ScanNet v2 are shown in Figure~\ref{fig:qualitative_scannet_supp}. 
The results on SUN RGB-D are shown in Figure~\ref{fig:qualitative_sunrgbd_supp}. 

\end{appendix}

{\small
\bibliographystyle{ieee_fullname}
\bibliography{egbib}
}

\end{document}